\documentclass[10pt,a4paper]{article}

\usepackage{caption}
\usepackage{subcaption}
\usepackage{graphicx}
\usepackage{wrapfig}
\graphicspath{{figures/}}

\usepackage{amssymb}
\usepackage{amsmath} 

\usepackage{algorithmic}
\usepackage{array}
\usepackage{lipsum}
\usepackage{pgfplots}
\usepackage{hyperref}
\usepackage{caption}
\usepackage{float}
\usepackage{tikz}
\usetikzlibrary{decorations}
\usepackage{rotating}

\usepgfplotslibrary{external}
\pgfplotsset{compat=1.18}
\begin{document}

\title{Byzantine-Robust Federated Learning: An Overview With Focus on Developing Sybil-based Attacks to Backdoor Augmented Secure Aggregation Protocols} 
\author{Atharv Deshmukh\\[0.5cm] Advisor: Suleyman Uludag}
\date{\today} 
\maketitle 

\begin{abstract}
\normalsize
Federated Learning (FL) paradigms enable large numbers of clients to collaboratively train Machine Learning models on private data. However, due to their multi-party nature, traditional FL schemes are left vulnerable to Byzantine attacks that attempt to hurt model performance by injecting malicious backdoors. A wide variety of prevention methods have been proposed to protect frameworks from such attacks. This paper provides a exhaustive and updated taxonomy of existing methods and frameworks, before zooming in and conducting an in-depth analysis of the strengths and weaknesses of the Robustness of Federated Learning (RoFL) protocol. From there, we propose two novel Sybil-based attacks that take advantage of vulnerabilities in RoFL. Finally, we conclude with comprehensive proposals for future testing, describe and detail implementation of the proposed attacks, and offer direction for improvements in the RoFL protocol as well as Byzantine-robust frameworks as a whole.

\end{abstract}

\setcounter{tocdepth}{3}
\tableofcontents

\clearpage

\section{Introduction}
\label{sec:introduction}

First introduced in February 2016 in the Google AI paper “Communication-Efficient Learning of Deep Networks from Decentralized Data”\cite{pmlr-v54-mcmahan17a}, Federated Learning (FL) and other distributed Machine Learning (ML) methods have quickly grown in both applications and popularity. These methods were initially introduced to enhance privacy and security by allowing deep training of complex algorithms with data from multiple sources, without publicizing and consolidating them. However, as the technology has evolved, it has become clear that FL comes with its own plethora of security and performance vulnerabilities. Due to the multi-party collaborative nature of FL protocols, systems have been left open to a flurry of attacks from multiple points, and are at risk if even a small percent of clients are compromised. 

This paper will provide a comprehensive overview of security and privacy in robust FL systems. An overview of this paper's contributions are shown as follows:
\begin{itemize}
\item Establish and expand an updated taxonomy of existing methods and frameworks for Byzantine-robustness in Federated Learning systems.
\item Provide a comprehensive evaluation of the strengths and weaknesses of the RoFL augmented secure aggregation protocol proposed by Burkhalter et al. \cite{lycklama2023rofl}.
\item Construct and propose two theoretical Sybil-based attacks to circumvent the RoFL framework based on these strengths and weaknesses.
\item Offer critical proposals for future implementation of this attack, outline future testing and analysis of the RoFL protocol, and give direction to further improvements in frameworks for Byzantine-robust FL
\end{itemize}

The rest of the paper is organized as follows: Section 2 introduces foundational concepts, starting with the architecture of a typical FL scheme and continuing by establishing a threat model and developing background about existing attack strategies. Section 3 offers a comprehensive and updated taxonomy of existing FL frameworks for Byzantine-robustness along with a detailed breakdown of the RoFL framework proposed by Burkhalter et al. Section 4 goes on to analyze the strengths, weaknesses, and vulnerabilities of the RoFL protocol, and Section 5 develops a novel theoretical attack proposal that exploits these vulnerabilities. Section 6 outlines possible avenues for future work in designing a system that guarantees Byzantine-robustness in FL paradigms, and finally, Section 7 offers a conclusion.

\section{Background}
\label{sec:background}

\textbf{Understanding FL.} In traditional ML paradigms, data from various sources is compiled in one centralized server and used to train complex models. Rather than bringing all data together, FL allows multiple clients to train their own models before sending only the models to a centralized server. This centralized server consolidates these updates using a process called Linear Aggregation which fundamentally averages out the parameters of the various clients to create a global model. In a process where there are $n$ clients and $w_i$ represents the model parameters of client $i$, the aggregated model parameters $w$ can be represented as: $w = (1/n) * \sum w_i$. This formula makes it so that updates are weighted based on size which, as discussed further in this paper, can be exploited by malicious entities. It is important to consider that due to large scale, most modern systems utilize Stochastic Gradient Descent \cite{pmlr-v54-mcmahan17a, 9154332}, selecting a random subset of client updates to aggregate each round. Following aggregation, the central server sends this global model back to clients, who further refine it and send it back to the central server in each subsequent round of training. This iterative process can be thought of as four cyclical steps: local training, model upload, aggregation, and model distribution. 

One of the many reasons for this extensive process is data privacy. Having a system in which no data, only model updates, are shared makes it possible and accessible to train complex and useful models while still satisfying the expectations of sensitive data privacy, regulatory compliance, and user trust. Take the often-used example of a group of hospitals working together to train a model based on private medical data. The very nature of sensitive patient data makes it illegal or unethical to publicize and thus impossible to use in a traditional ML paradigm. However, with the advent of FL, hospitals are able to collaborate effectively  in a privacy preserving manner, training models on their own private servers and sending only their model updates to a federated server. 

\begin{figure}
    \centering
    \includegraphics[width=1.0\linewidth]{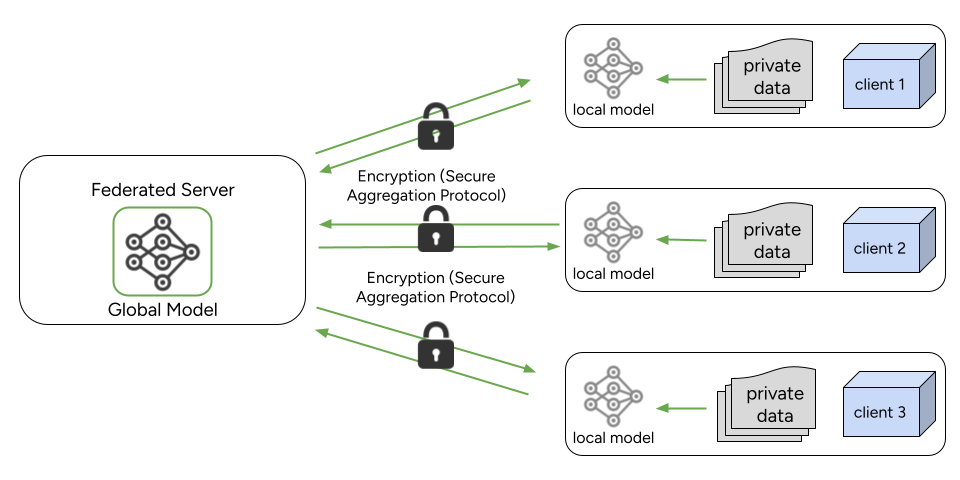}
    \caption{Architecture of a typical FL setup with a Secure Aggregation protocol to protect client updates from inference attacks.}
    \label{fig:architecture}
\end{figure}

\vspace{1\baselineskip}
\noindent\textbf{Threat Model.} In the context of privacy and security, two aspects stand out as paramount: protecting data security and preserving model performance. Due to the inherent tendency of FL systems to involve sensitive data, it follows logically that bad actors will attempt to gain access to said data. Apart from simply hacking the client machines, by far the most prevalent of systemic vulnerabilities to such attacks is the previously defined process of model uploading. By tracking specific model updates and weights from clients, it becomes possible for third parties to deduce the approximate or exact sensitive data that clients are using to train their models. For such inference attacks \cite{shokri2017membershipinferenceattacksmachine, carlini2021extractingtrainingdatalarge}, FL protocols have evolved to integrate Secure Aggregation. At a fundamental level, Secure Aggregation protocols encrypt model updates before sending them to the federated server and decrypt them before aggregation. This encryption allows for an extra layer of defense between sensitive data and attackers. While it effectively defends against inference attacks, makes it difficult to examine or audit clients’ model updates or data.

Another less developed but just as significant area of concern in FL systems is model integrity and protection. Oftentimes, FL frameworks are susceptible to attacks that attempt to change the behavior or hurt the performance of the global model. Due to their multi-stakeholder nature, having even a small number of compromised clients can significantly alter and hurt the overall performance of the model; i.e. linear aggregation is not Byzantine resilient \cite{NIPS2017_f4b9ec30}. Malicious third-parties use a variety of attack strategies and methods, broadly referred to as Byzantine attacks \cite{10.1145/357172.357176}, to hijack clients, target specific areas of function, and compromise model function. This paper will focus on analyzing threats to model integrity in FL protocols, testing and evaluation of existing solutions, and proposing ideas for future advancements of these frameworks.

\vspace{1\baselineskip}
\noindent\textbf{Defining Robustness.} Before analyzing model integrity, it is essential to define a measurement for the specific characteristics we want to evaluate. Considering that our aim is to evaluate our system’s ability to withstand attacks from a malicious third party and maintain a high level of performance, the metric best suited to our purposes is robustness, or more specifically, adversarial robustness. In our context, if a system is able to maintain a high level of accuracy and consistency in the presence of adversaries, it is considered robust.

\begin{figure}
    \centering
    \includegraphics[width=.75\linewidth]{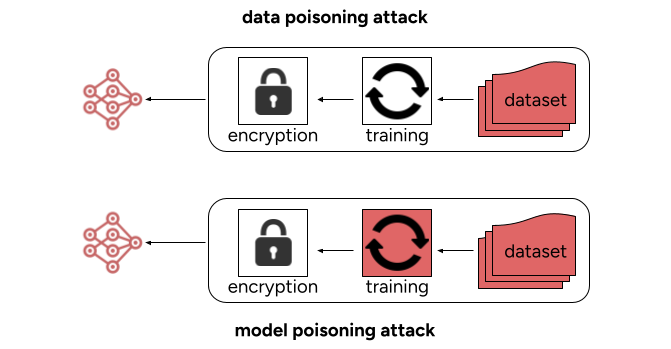}
    \caption{Data poisoning attacks versus model poisoning attacks. Figure inspired by \cite{lycklama2023rofl}.}
    \label{fig:poisoning-attacks}
\end{figure}

\vspace{1\baselineskip}
\noindent\textbf{Understanding Attack Strategies.} As mentioned before, when targeting FL systems, attackers utilize a wide range of methods with the aim of influencing and gaining control of the global model. Developing a basic understanding of these methods is crucial before we evaluate their effectiveness and engineer robust systems to limit or completely prevent them. Attacks on model integrity generally fall into two categories: data poisoning and model poisoning.

True to the name, data poisoning simply involves feeding the model incorrect or mislabeled data in order to make it misbehave. An example of data poisoning would be feeding a disease detection model pictures of pneumonia wrongly labeled as lung cancer. Specifically in the case of FL systems, due to the nature of the nearly ubiquitous linear aggregation methodology of consolidating model updates, having even a few clients affected by poisoned data can degrade or corrupt the global model. Although as will be discussed, certain factors may cause data poisoning attacks to be more or less effective. It is also worth mentioning that data poisoning itself does not exploit any specific vulnerabilities in FL systems, and instead stands as a common attack strategy across different types of ML paradigms.

Model poisoning attacks center around hijacking client machines and exploiting FL protocols by sending damaging updates straight to the federated server. As the central server aggregates these updates, the global model gets corrupted and drops in performance. Model poisoning has been shown to generally be a more effective \cite{pmlr-v108-bagdasaryan20a} attack strategy than data poisoning, and is therefore more heavily relied upon in modern attacks. It is also worth mentioning that model poisoning attacks are either targeted or untargeted. Targeted attacks send in model updates that focus on a specific aspect of functionality, while untargeted attacks aim to simply hurt the overall functionality of the model.

Model poisoning and data poisoning attacks can be classified as either single-shot or continuous. As the names suggest, single-shot attacks are introduced and take place only in one round of training, whereas continuous attacks involve corrupted updates being sent in for multiple successive rounds. 

Another attack strategy almost ubiquitous in poisoning attacks is scaling. After taking over client machines, adversaries generate corrupted updates that are much larger in scale than updates coming from honest clients. As discussed earlier, the linear aggregation system used in FL protocols simply puts together client updates, weighting them based on size. By virtue of their large size, these corrupted updates have an outsized impact on model performance. Various strategies to combat the impact of scaling will be discussed further.

\section{Related Work}
\label{sec:related-work}
 Before delving into an evaluation of RoFL and proposals for future work, it is necessary to establish an understanding of existing methods for FL robustness. We will start off at a basic level, examining various proposed strategies. We will then move on to building an understanding of the RoFL secure aggregation protocol proposed by Burkhalter et al.\cite{lycklama2023rofl}
 with the aim of contextualizing the proposals that will be made in this paper.

\begin{table}[ht!]
\begin{tabular}{ |p{1.75cm}||p{1.75cm}|p{1.75cm}|p{1.75cm}|p{1.75cm}|p{1.75cm}|p{1.75cm}| }
 \hline
 \multicolumn{7}{|c|}{\textbf{Existing Robust FL Frameworks}} \\
 \hline
 \textbf{Solution}&\textbf{Category}&\textbf{Target Attack}&\textbf{Max Attackers}&\textbf{Model Accuracy}&\textbf{Data Distribution}&\textbf{Time Complexity}\\
 \hline
 Multi-Krum \cite{NIPS2017_f4b9ec30}&Distance based&Data/model poisoning&50\%&Medium&IID&$O(K^2d)$\\
 \hline
 FABA \cite{unknown(1)}&Distance based&Data/model poisoning&50\%&Medium&IID&$O(K^2d)$\\
 \hline
 Sniper \cite{Cao2019UnderstandingDP}&Distance based&Data poisoning&50\%&Medium&IID&$O(K^2d)$\\
 \hline
 FoolsGold \cite{DBLP:journals/corr/abs-1808-04866}&Distance based&Data/model poisoning&No limit&Medium High (with Multi-Krum)&IID/non-IID&$O(K^2d)$\\
 \hline
 Wan et al. \cite{inproceedings}&Distance based&Model Poisoning&50\%&High&IID&$O(K^2d)$\\
 \hline
 MAB-RFL \cite{Wan_2022}& Distance based&Data/model poisoning&50\%&High&IID/non-IID&$O(K^2d)$\\
 \hline
 Li et al. \cite{unknown(2)}&Performance based&Data/model poisoning&50\%&High&IID/non-IID&$O(Kd)$\\
 \hline
 Zeno \cite{pmlr-v97-xie19b}&Performance based&Data/model poisoning&No limit&High&IID/non-IID&$O(Kd)$\\
 \hline
 Cao et al. \cite{8861393}&Performance based&Data/model poisoning&No limit&High&IID&$O(Kd)$\\
 \hline
 FLTrust \cite{cao2022fltrustbyzantinerobustfederatedlearning}&Performance based&Data/model poisoning&No limit&High&IID/non-IID&$O(Kd)$\\
 \hline
 AFA \cite{muñozgonzález2019byzantinerobustfederatedmachinelearning}&Statistic based&Data/model poisoning&50\%&High&IID&$O(K^2d)$\\
 \hline
 GeoMed MarMed Trimmedmean \cite{xie2018generalizedbyzantinetolerantsgd, yin2021byzantinerobustdistributedlearningoptimal}&Statistic based &Data/model poisoning&50\%&High&IID/non-IID&$O(KdlogK)$\\
 \hline
 Bulyan \cite{mhamdi2018hiddenvulnerabilitydistributedlearning}&Statistic based&Data/model poisoning&50\%&Medium&IID&$O(K^2d)$\\
 \hline
\end{tabular}
\caption{(Part 1) Comprehensive taxonomy of existing robust FL frameworks \cite{shi2022challengesapproachesmitigatingbyzantine}. Model Accuracy represents the prediction accuracy of the scheme, "Medium" and "High" indicate the accuracy is below and close to the non-attacker case respectively. IDD means the dataset is identically and independently distributed and non-IDD indicates it is not. $K$ denotes the number of users and $d$ denotes the model size. Adapted and modified from Shi et al. (2022) [arXiv:2112.14468v2].}
\label{table:1}
\end{table}

\begin{table}[ht!]
\begin{tabular}{ |p{1.75cm}||p{1.75cm}|p{1.75cm}|p{1.75cm}|p{1.75cm}|p{1.75cm}|p{1.75cm}| }
 \hline
 \multicolumn{7}{|c|}{\textbf{Existing Robust FL Frameworks}} \\
 \hline
 \textbf{Solution}&\textbf{Category}&\textbf{Target Attack}&\textbf{Max Attackers}&\textbf{Model Accuracy}&\textbf{Data Distribution}&\textbf{Time Complexity}\\
 \hline
  SLSGD \cite{xie2019slsgdsecureefficientdistributed}&Statistic based&Data/model poisoning&50\%&Medium&IID/non-IID&$O(KdlogK)$\\
 \hline
 RFA \cite{Pillutla_2022}&Statistic based&Data/model poisoning&50\%&Medium&IID&$O(Kd)$\\
 \hline
 RSA \cite{li2019rsabyzantinerobuststochasticaggregation}&Target optimization based&Data poisoning&No limit&High&IID/non-IID&$O(Kd)$\\
 \hline
 NoV \cite{xing2024vandalismprivacypreservingbyzantinerobustfederated}&Distance based&Data/model poisoning&50\%&High&IID/non-IID&$O(K^2d)$\\
 \hline
 DisBezant \cite{article}&Distance based&Data/model poisoning&50\%&High&IID/non-IID&$O(K^2d)$\\
 \hline
 DRACO \cite{chen2018dracobyzantineresilientdistributedtraining}&Statistic Based&Data/Model poisoning&50\%&High&IID/non-IID&$O(K^2n)$\\
 \hline
 Auror \cite{10.1145/2991079.2991125}&Distance based&Data/Model poisoning&no limit&High&IID/non-IID&$O(K^2n)$\\
 \hline
 BPFL \cite{nie2024efficientbyzantinerobustprovablyprivacypreserving}&Distance Based&Data/Model poisoning&50\%&High&IID/non-IID&$O(K^2n)$\\
 \hline
 RoFL \cite{lycklama2023rofl}&Statistic based&Data/model poisoning&50\%&High&IID/non-IID&$O(K^2d)$\\
 \hline
 \end{tabular}
\caption{(Part 2) Comprehensive taxonomy of existing robust FL frameworks \cite{shi2022challengesapproachesmitigatingbyzantine}. Model Accuracy represents the prediction accuracy of the scheme, "Medium" and "High" indicate the accuracy is below and close to the non-attacker case respectively. IDD means the dataset is identically and independently distributed and non-IDD indicates it is not. $K$ denotes the number of users and $d$ denotes the model size. Adapted and modified from Shi et al. (2022) [arXiv:2112.14468v2].}
\label{table:2}
\end{table}

\vspace{1\baselineskip}
\noindent\textbf{Taxonomy of Existing Proposals.} As defined by Shi et al. in “Challenges and Approaches for Mitigating Byzantine Attacks in Federated Learning” \cite{shi2022challengesapproachesmitigatingbyzantine}, existing defenses against Byzantine attackers can be divided into four categories based on their methodologies: distance based, performance based, statistics based, and target optimization based. Distance based defense mechanisms differentiate and discard updates based on how far they are from the others; performance based frameworks test each update on a clean, server-provided dataset and either decrease weights of or automatically discard updates that perform poorly; statistics based frameworks utilize the statistical properties of the updates and broadly discard ones that fall far from the mean, median, or other chosen measure; and target optimization based schemes optimize a different objective function in order to improve the overall robustness of the model. For the purposes of this paper, we will use the taxonomy put together by Shi et al., modified slightly to include categorization more relevant to our purposes, and with an addendum to include frameworks proposed after late 2022.

\vspace{1\baselineskip}
\noindent\textbf{Norm Bounds.} As discussed earlier, a large number of poisoning attacks rely on scaling \cite{pmlr-v97-bhagoji19a, pmlr-v108-bagdasaryan20a} to have a greater impact. Making malicious updates magnitudes larger than honest updates allows adversaries to exert significant influence over the global model with a small number of compromised clients. In order to combat this, robust FL systems have evolved to include Norm Bounds, which at a mathematical level simply set an upper limit on the magnitude of a vector, matrix, or operator. In our context, norm bounds place a cap on the $p$-norm, or size of updates coming from individual client machines, thereby nullifying the impact of scaling and limiting malicious clients’ effects on the global model.

Burkhalter et al. utilize an experimental setup with a digit classification and image classification model, each with a constant number n of m compromised clients (represented as $\alpha \in [0, 1])$ per training round, and focuses on two main metrics: main task accuracy and backdoor task accuracy. Main task accuracy shows the performance of the model on the intended task while backdoor accuracy represents the model’s performance on a selected malicious task. Using these methodologies, it is shown that norm bounds, when implemented in a manner where they are not excessively tight or loose, serve as an effective defense against a variety of single-shot and continuous model poisoning attacks. However, in scenarios where the adversary controls a larger number of clients or specifically chooses tail targets \cite{wang2020attacktailsyesreally, NEURIPS2020_b8ffa41d}, which are subpopulations that are underrepresented in training data, norm bounds are not as effective in preventing the effects of Byzantine participants.

\begin{figure}
    \centering
    \includegraphics[width=1.0\linewidth]{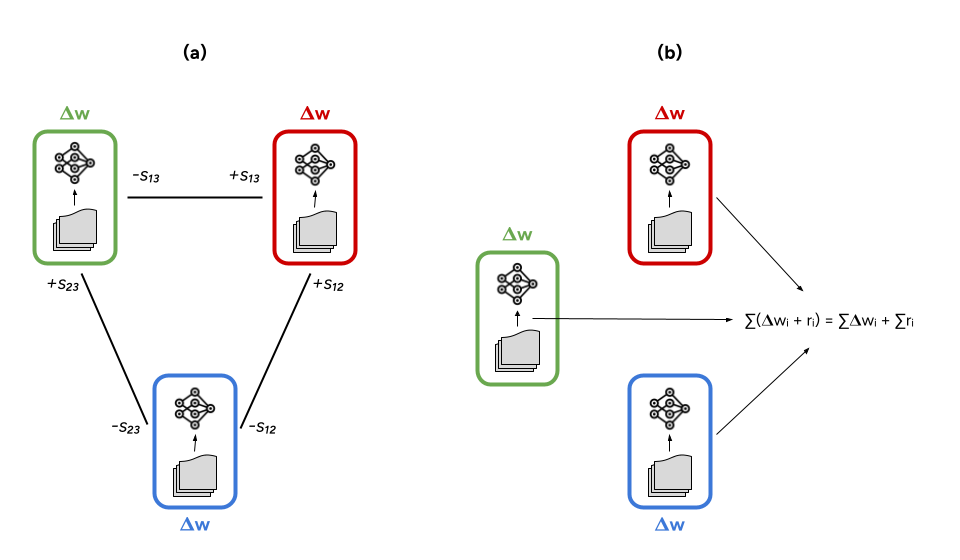}
    \caption{Additive homomorphic scheme proposed by Bonawitz et al. \cite{10.1145/3133956.3133982} (a) Additive masks based on pairwise shared secrets $s_{ij}$. $r_1 + r_2 + r_3 = 0$ where $r_1 = s_{12} + s_{13}$, $r_2 = -s_{12} + s_{23}$, and $r_3 = -s_{13} - s_{23}$. (b) Server adds the updates using the homomorphic properties of the scheme. Figure inspired by \cite{lycklama2023rofl}.}
    \label{fig:homomorphic-masking}
\end{figure}

\vspace{1\baselineskip}
\noindent\textbf{Understanding RoFL.} In the context of other existing methods to promote FL robustness, norm bounds \cite{Sun2019CanYR} stand out as a promising approach. However, there still exist some challenges in the way of implementing norm bounds effectively. The privacy preserving nature of Secure Aggregation protocols makes it difficult to enforce norm bounds, and with actively malicious clients, machines cannot be trusted to self-regulate \cite{lycklama2023rofl}. The RoFL framework aims to modify existing Secure Aggregation protocols to enforce norm bounds over private updates from untrusted clients without being too impractical to implement in practice.

RoFL is built upon the additive homomorphic (meaning operations can be performed without decrypting data) masking based secure aggregation protocol proposed by Bonawitz et al. \cite{10.1145/3133956.3133982} and later extended by Bell et al. \cite{Bell2020SecureSA}. At a basic level, this additive masking scheme involves each client machine obscuring it's input vector $w$ with a masking vector $r$. These masking vectors then cancel out in the aggregation process, allowing the server to recover the sum. The protocol can be broken down into three stages. During mask distribution, clients agree on the cryptographic share keys used to generate the mask. This is followed by aggregation, in which the clients send their masked updates to the server, allowing the server to take advantage of the additive properties of the scheme and combine the updates as such: $\sum(\Delta w_i + r_i) = \sum \Delta w_i + \sum r_i$. Finally, the federated server receives a decoding key $r'$ that allows it to cancel out the masking values of clients, leaving it with only the updates: $(\sum \Delta w_i + \sum r_i) - r' = \sum \Delta w_i$. Although this secure aggregation protocol does provide protection against inference attacks, it does not address the need for possibly malicious clients to prove correctness of their updates. 

Proving correctness would involve clients having to prove that they masked with the correct value, which in turn, would mean proving they correctly followed the share keys protocol. To avoid this complicated process with an expensive overhead, Burkhalter et al. show that the only proof needed is that the sum of the masks, $\sum r_i$ is equal to the decoding key $r'$, and propose a novel encoding scheme to allow the server to verify the sum in an efficient manner by taking advantage of the homomorphic properties of the framework. 

When choosing an appropriate Zero Knowledge Proof (ZKP) for proving correctness, it is necessary to take into account overhead, prover time, and fit into homomorphic schemes among other factors. With these factors in mind, bulletproofs \cite{8418611}, with a proof size of $O(log(\ell))$, prover time of $O(\ell)$, and verification time of $O(\ell)$, along with the ability to operate directly on homomorphic commitments, stand out as the best option. Bulletproofs are traditionally implemented on top of Pederson \cite{10.1007/3-540-46766-1_9} commitments, which utilize the discrete logarithm problem in order to seal a value without revealing it. A Pederson commitment in the context of RoFL would take the form of $C = g^{\Delta w_i} h^r_i$, with $C$ being the commitment, and $g$ and $h$ being generator elements. In order to allow the server to homomorphically add up check the value of $r'$, this Pederson commitment is expanded to include a term of $g^r_i$, letting the server compare $\sum g^r_i$ to $g^{r'}$. In addition to these commitments, RoFL requires each client to send a Bulletproof-based ZKP that the sever verifies in order to check that the update satisfies the norm bound. With further optimizations, the timing overhead of RoFL decreases and as parameters increase, it decreases asymptotically. 
\section{Analysis}
\label{sec:analysis}
In order to set the groundwork for an adversarial attack that can successfully bypass RoFL, it is necessary to conduct an in-depth analysis on the protocol's strengths and weaknesses. This theoretical assessment will draw from various evaluations of components of RoFL as well as the the original paper.

\vspace{1\baselineskip}
\noindent\textbf{Norm Bounds Vulnerabilities.} When examining the protocol, the most obvious vulnerability comes from RoFL's reliance on norm bounds. As discussed earlier, norm bounds, although great for reducing overhead and efficiency, prove not to be exhaustive in terms of preventing all types of attacks. From experimentation carried out by Burkhalter et al., we can see that norm bounds offer effective protection against single-shot attackers and continuous attackers on prototypical targets. However, when faced with a continuous attack on a tail target, norm bounds fail to protect model integrity. It is important to consider that this kind of attack requires at least one malicious client to be selected in a most if not all rounds, which can be achieved in two ways: either probabilistically by simply having enough adversaries in random sampling attacks, or through fixed frequency attacks where there can be one adversary selected in each round \cite{pmlr-v108-bagdasaryan20a}. 

Reliance on norm bounds in the interest of decreasing overhead also makes it so that the protocol has no control over whether updates are malicious or not. Rather, norm bounds simply ensure that as long as there are more honest clients than adversarial ones, bad updates are overpowered. As the number of malicious clients increases, scaling becomes less relevant for attack success and poisoning attacks become more likely to succeed \cite{lycklama2023rofl}.

Another possible vulnerability in RoFL stems from the actual calculation of the norm bound. As shown by Burkhalter et al., the norm bound being too tight or loose will either slow down training drastically or make way for malicious updates to influence the global model, respectively \cite{lycklama2023rofl}. Since the norm bound is dynamically calculated using the median update size and each client's update size is made public, significant adversarial presence will allow malicious clients to manipulate the norm bound as desired. 

\vspace{1\baselineskip}
\noindent\textbf{Collaborative Attacks.} One dimension that RoFL does not directly address is threat of joint attacks, or attacks based on collusion between multiple malicious clients. Relative ease of collusion enables adversaries to more easily collaborate to achieve malicious goals such as manipulating the median, selecting tail targets, or conducting fixed frequency attacks. 

In addition to not preventing collaborative attacks, RoFL also does not protect against the threat of falsely introducing new clients. Once again, having a seemingly larger set of clients makes focusing on tail targets and manipulating statistical measures significantly easier. This class of attacks, broadly known as Sybil attacks \cite{9767718}, will be further discussed in following sections.

\vspace{1\baselineskip} 
\noindent As established earlier and in Burkhalter et al.'s original paper, RoFL stands as an effective method to prevent a wide range of practical poisoning attacks on FL schemes. The augmented secure aggregation protocol also features relatively low overhead as well as several optimizations that work to reduce time cost. This focus on practicality comes at a cost in the context of certain attacks, namely attacks on tail targets, and gives room to joint and Sybil attacks. Note that instructions for running and testing the RoFL protocol along with related analyses is included in the Appendix.
\section{Proposed Attack}
\label{sec:proposed-attack}
From an adversarial point of view, tail targeted attacks and manipulation of statistics seem to be the most promising venues to infiltrate the protocol and compromise model integrity. Looking at these two approaches, it becomes clear that one possible way to exploit them both would be by simply having more malicious clients. With an increased number of adversaries poisoning attacks take hold at a faster rate \cite{article} and it becomes more possible to manipulate the median to loosen the norm bound to a desired extent. 

\begin{figure}
    \centering
    \includegraphics[width=1.0\linewidth]{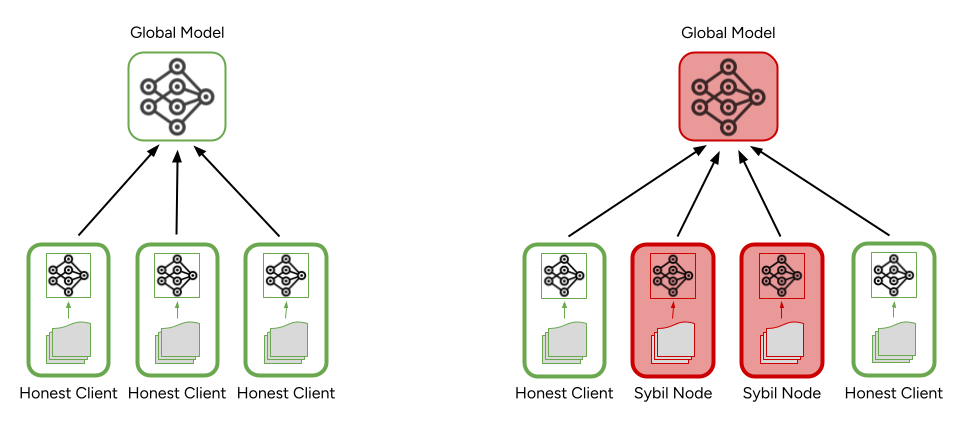}
    \caption{FL system with and without colluding Sybil nodes mounting a poisoning attack. Sybil-based attacks involve adversaries creating multiple fake identities or Sybil nodes \cite{9767718}.}
    \label{fig:sybil-nodes}
\end{figure}

\vspace{1\baselineskip} 
\noindent\textbf{Sybil-based Attacks.} As discussed earlier, Sybil-based attacks, first proposed by Xiao et al. \cite{9767718}, involve adversaries creating multiple fake identities or "Sybil nodes" in order to amplify their impacts. This is enabled by the fact that a core property of most FL aggregation systems is support for participants to join and leave periodically. In the attack described by Xiao et al., each Sybil node contains a cloned poisoning model. This effectively has the same effect as scaling in the absence of norm bounds, causing large adverse impacts on the global model. This is then taken further with collusion between the Sybil nodes in order to avoid triggering anomaly detection frameworks. Due to the large number of nodes, there is also a larger probability of a malicious client being chosen in each training round. In the context of circumventing RoFL, this attack framework adequately provides the tools necessary to both statistically manipulate the median norm and more effectively perpetuate tail targeted attacks. The following subsections will outline possible extensions of this attack to further exploit the RoFL protocol.

\vspace{1\baselineskip} 
\noindent\textbf{Sybil-Tail Attacks.} First proposed by Wang et al. \cite{NEURIPS2020_b8ffa41d}, tail attacks stand out as continuing to be effective under norm bounds. As discussed by Burkhalter et al., the main reason for this is because attacks on tail targets take advantage of the learning capacity of the model; there are simply not enough honest data points to outweigh the effect of malicious updates. Using an approach with Sybil nodes, it is possible to accelerate and strengthen attacks on tail targets. A larger number of clients also naturally translates to having a larger dataset and consequentially, a larger number of tail targets. These newly created "artificial tail targets" can possibly help provide increased surface for attacks, further compromising the system. 

In addition to tail targeted attacks, Sybil attacks once again open the door to other kinds of poisoning attacks. Although scaling is effectively prevented by RoFL, having multiple Sybil-nodes can allow bad actors to dominate aggregation in much the same way. 

\vspace{1\baselineskip} 
\noindent\textbf{Sybil Attacks for Statistical Manipulation.} Another possible direction for backdooring RoFL is through manipulation of the median. As discussed before, the norm bound is defined by multiplying the statistical median by an arbitrary small multiple. It is worth mentioning that in this process, the update sizes of each client are made public. When the majority of clients are honest, this process ensures that incoming updates are relatively safe without slowing down training. With ordinary attack frameworks, statistical manipulation of this sort requires a malicious entity to control close to $50\%$ of clients. Sybil-based attacks however, allow one single client to create arbitrarily many fake identities \cite{9767718}. If each one of these fake identities sends updates that are larger than average, the norm bound will consequently loosen. 

When thinking about statistical manipulation of this sort, it is important to consider that there will be measurable performance losses on the global due to aggregation dominating, long before there are a sufficient enough number of malicious clients to create these effects. Rather than standalone effects, it is likely that statistical manipulation will result in an acceleration of model degradation as more Sybil nodes get created and the norm bound continues to loosen.
\section{Future Work}
\label{sec:future-work}
Future steps that most logically follow include further experimental testing and analysis of RoFL, implementation and assessment of the proposed Sybil-Tail and Statistical Manipulation attacks, and eventually, design of increasingly strong protocols for Byzantine-robustness. In order to both outline the next steps of this project and offer direction to other researchers, each will be outlined below. 

\vspace{1\baselineskip} 
\noindent\textbf{Testing and Analysis.} Although the scope of this paper is largely theoretical and draws effectively from the experimentation of other researchers, further testing of the RoFL protocol could prove useful. Possible directions include testing the framework in the presence of statistical manipulation attacks or testing with existing Sybil-based poisoning attacks as described by Xiao et al. \cite{9767718}. Results from this testing could be used both to find new vulnerabilities and vet the practicality of Sybil-based attacks on RoFL. Details for testing and analyzing the protocol, including original code, are included in the Appendix.

\vspace{1\baselineskip} 
\noindent\textbf{Attack Implementation.} In order to test their effectiveness and eventually build stronger protocols for Byzantine-robustness, it is paramount that the two Sybil-based attacks proposed in this paper be implemented. Similar to other proposed attacks \cite{9308910}, implementation would take the form of developing a concrete simulation or prototype and evaluating various benchmarks such as backdoor task accuracy, training time, and overhead among others. 

\vspace{1\baselineskip} 
\noindent\textbf{Furthering Byzantine-robust Protocols.} In order to improve protections against Sybil-based attacks like the ones proposed, frameworks can evolve to include distance based anomaly detection as found in FoolsGold proposed by Fung et al. \cite{DBLP:journals/corr/abs-1808-04866}, while keeping the largely practical benefits offered by methods like RoFL. Altogether other avenues include exploring protocols built on the use of differential privacy to add noise to client updates \cite{10.1007/978-3-031-07689-3_33, unknown} and building augmented secure aggregation protocols based on different secure aggregation schemes.
\section{Conclusion}
\label{sec:conclusion} 
Federated Learning systems allow the distributed training of complex models on private data across a large number of machines. They have an immense number of applications across areas ranging from medical technology to large networks of mobile phones. But the very nature of FL opens these schemes up to attacks that either try to steal sensitive and private data by spying on client updates, or degrade the overall performance of the model by hijacking clients and sending corrupted updates. In order to address the former, systems have evolved to include secure aggregation protocols which encrypt client updates. Unfortunately, this extra level of privacy makes it even more difficult to ensure that clients are sending honest updates. Over the years, dozens of protocols have been proposed with the aim of minimizing or altogether preventing these poisoning attacks.

This paper provides a comprehensive and updated taxonomy of proposed frameworks to protect the integrity of FL systems. After establishing the strengths, weaknesses, and goals of each framework, we focus on breaking down the vulnerabilities of the RoFL secure aggregation protocol proposed by Burkhalter et al. Using this analysis, we construct two novel Sybil-based attacks to circumvent the RoFL protocol. Finally, we conclude with proposals for future work both specifically related to RoFL and systems for Byzantine-robustness in general.

As distributed networks of clients increasingly desire to collaborate on building ML models in a privacy preserving manner, developing strategies for secure and Byzantine-robust FL becomes increasingly important. Additionally, as protective technology becomes stronger, the same happens with adversaries. In this ever-evolving landscape, designing new protocols as well as analyzing them from an adversarial angle is undoubtedly essential to take Federated Learning systems to their fullest potential. 

\clearpage

\appendix
\section{Appendix}
\label{sec:appendix}
In order to conduct further empirical analysis and build on top of Burkhalter et al.'s RoFL protocol, it is necessary to replicate and run the prototype along with related analysis code. Code for the testing framework can be found at \url{https://github.com/pps-lab/fl-analysis} and code for the implemented protocol can be found at \url{https://github.com/pps-lab/rofl-project-code}. Since running this code may prove difficult across various machines and operating systems, this appendix will discuss some important pointers to navigate this process.

\vspace{1\baselineskip}
\noindent\textbf{Installation Process.} Both the analysis framework and implemented protocol were developed and originally run on Ubuntu 18.04, and in order to minimize scope for version and Operating System related issues, it is recommended to us the same environment when testing. Depending on the system, the process for getting Ubuntu 18.04 running can differ but it is almost always easiest to run a Virtual Machine (VM) using either VirtualBox (for Windows and older Macbooks) or UTM (for newer Macbooks). After setting up a VM and going through all arbitrary processes, following the instructions given in the GitHub proves to be sufficient. For any further questions or concerns regarding testing, contact information for the original authors is linked below.

\vspace{1\baselineskip}
\noindent Lukas Burkhalter: \href{mailto:lukas.burkhalter@lubu.info}{lukas.burkhalter@lubu.info}

\noindent Hidde Lycklama: \href{mailto:hidde.lycklama@inf.ethz.ch}{hidde.lycklama@inf.ethz.ch}

\clearpage
\bibliography{bibliography/references}{}
\bibliographystyle{IEEEtran}

\end{document}